\pdfoutput=1

\documentclass{article}

\usepackage{microtype}
\usepackage{graphicx}
\usepackage{booktabs}
\usepackage{array}
\usepackage{multirow}
\usepackage{amsmath,amssymb}
\usepackage{xspace}
\usepackage{xcolor}
\usepackage{tikz}
\usepackage{enumitem}
\usepackage{hyperref}
\usepackage{xurl}
\usepackage{needspace}

\usepackage[accepted]{mlsys2025}
\makeatletter
\renewcommand{\Notice@String}{}
\renewcommand{\printAffiliationsAndNotice}[1]{%
  \stepcounter{@affiliationcounter}%
  {\let\thefootnote\relax\footnotetext{\hspace*{-\footnotesep}%
    \forloop{@affilnum}{1}{\value{@affilnum} < \value{@affiliationcounter}}{%
      \textsuperscript{\arabic{@affilnum}}\csname @affilname\the@affilnum\endcsname
      \ifnum\value{@affilnum}<\numexpr\value{@affiliationcounter}-1\relax\space\fi}.}}}
\makeatother
\hypersetup{
  hypertexnames=false,%
  pdftitle={Detent: Batch-Invariant LLM Serving and RL Rollouts at Near-Stock Speed},
  pdfsubject={},
  pdfkeywords={batch invariance, deterministic inference, LLM serving, reinforcement learning, kernel selection, autotuning}}

\newcommand{\sys}{\textsc{Detent}\xspace}
\newcommand{\selir}{SelIR\xspace}
\newcommand{\sgdet}{SGLang-Det\xspace}
\newcommand{\vldet}{vLLM-Det\xspace}
\newcommand{\tx}{\ensuremath{\times}\xspace}
\newcommand{\code}[1]{\texttt{#1}}

\usetikzlibrary{positioning,fit,backgrounds,arrows.meta,shapes.geometric,calc,decorations.pathreplacing}
\definecolor{rtred}{HTML}{C2412D}
\definecolor{okgreen}{HTML}{1D6A47}
\newcommand{\ok}{\textcolor{okgreen}{\ensuremath{\checkmark}}}
\newcommand{\no}{\textcolor{rtred}{\ensuremath{\times}}}
\makeatletter
\g@addto@macro\@floatboxreset{\normalcolor}
\makeatother



\mlsystitlerunning{Detent: Batch-Invariant LLM Serving and RL Rollouts at Near-Stock Speed}

\begin{document}

\twocolumn[
\mlsystitle{Detent: Batch-Invariant LLM Serving and RL Rollouts\texorpdfstring{\\}{ }at Near-Stock Speed}

\begin{mlsysauthorlist}
\mlsysauthor{Peixu Xin}{bu}
\mlsysauthor{Runchuan Feng}{bu}
\mlsysauthor{Jiazhen Guo}{google}
\mlsysauthor{Kan Wu}{radixark}
\mlsysauthor{Ming Yin}{princeton,gatech}
\end{mlsysauthorlist}
\mlsysaffiliation{bu}{Independent Researcher, Boston University, Boston, MA, USA}
\mlsysaffiliation{google}{Google, Mountain View, CA, USA}
\mlsysaffiliation{radixark}{RadixArk, Palo Alto, CA, USA}
\mlsysaffiliation{princeton}{Princeton University, Princeton, NJ, USA}
\mlsysaffiliation{gatech}{Georgia Institute of Technology, Atlanta, GA, USA}

\mlsyskeywords{batch invariance, deterministic inference, LLM serving, reinforcement learning, kernel selection, autotuning}

\vskip 0.3in

\begin{abstract}
Reproducible evaluation, verifiable inference and on-policy reinforcement learning (RL) need an LLM engine to give a request the same output bits whichever requests share its batch.
Batch-invariant kernels exist, but engines still select each kernel and its parameters from the batch.
Two kernels that compute the same sum can round it differently.
The engines' deterministic modes remove the choice by keeping one conservative kernel per operator and pay for it in latency.
Many fast kernels are batch-invariant by themselves, so it is enough to fix the few batch-dependent values that kernel selection reads.
\sys expresses kernel selection as a program that the engine executes.
It finds the fewest values in this program that must be fixed and tunes them for the workload before deployment.
On multi-GPU deployments of Llama-3.1-70B and Qwen3-30B-A3B in vLLM and SGLang, \sys is batch-invariant.
Tuned per load, it serves at 1.04--1.17\tx the latency per output token of Stock (each engine with its standard settings), against 1.67--10.3\tx for the deterministic modes.
In SGLang it matches the GSM8K accuracy of SGLang's deterministic mode.
\sys covers multiple deployment combinations, including multiple engines, GPU generations, tensor-parallel degrees, and dense and MoE models.
Seeded RL rollouts of models up to Llama-3.1-70B replay bit for bit under other batches and cache splits.
A trainer on \sys's kernels and values matches every rollout logprob bit for bit at 1.03\tx Stock's rollout time per token.
BF16 GRPO then stays stable where Stock's run collapses.

\end{abstract}
]

\printAffiliationsAndNotice{}

\section{Introduction}

A request is batch-invariant when its output bits do not depend on the other requests in its batch.
Engines with their standard settings, which we call Stock, are not.
Changing only the load changes Stock SGLang's answer on 33 of GSM8K's 1,319 problems (\S\ref{sec:eval-acc}).
A trainer that recomputes its rollouts matches only 1.0--47.7\% of their tokens bit for bit.
This mismatch alone can collapse RL training (\citealp{zhong2026vexact}; Figure~\ref{fig:rl}a,c).
Evaluations are not repeatable~\citep{yuan2025nondet}, and third parties cannot verify inference by re-execution~\citep{eigenai2026}.
Prior work makes the reduction order inside each kernel independent of the batch~\citep{thinkingmachines2025,tbik2026}.
The engine, however, still selects each kernel and its parameters from batch-dependent quantities such as the batch's token count.
Two kernels that compute the same sum can round it differently.

\begin{figure}[t]
\centering
\includegraphics[width=\columnwidth,trim=0 1.8pt 0 3.9pt,clip]{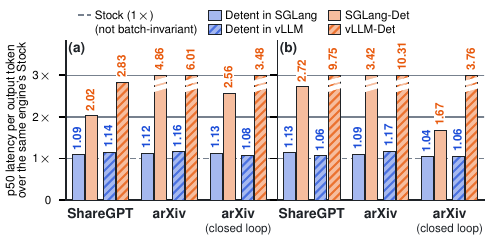}%
\caption{ShareGPT replays multi-turn chats. Their prompts are short and mostly hit the prefix cache, so most of the work is decode. arXiv sends 8k-token documents with Poisson arrivals and long prefills. Both run at loads up to Stock's peak. Closed loop sends the arXiv documents one or eight at a time, without queueing (\S\ref{sec:eval-setup}). (a)~Qwen3-30B-A3B; (b)~Llama-3.1-70B. Bars: geometric mean over a workload's loads, with \sys tuned per load (\S\ref{sec:tuner}).}
\label{fig:headline}
\end{figure}

SGLang~\citep{zheng2024sglang} removes the choice in its deterministic mode (\sgdet; \citealp{sglangdet2025}).
vLLM~\citep{kwon2023vllm} does the same in its batch-invariant mode (\vldet; \citealp{vllmbi2026}).
Both modes keep one conservative kernel per operator: unsplit attention, NCCL's tree all-reduce and one hard-coded MoE configuration.
They are batch-invariant, but their serving latency per output token is 1.59--6.34\tx (\sgdet) and 2.83--10.3\tx (\vldet) Stock's per load (Figures~\ref{fig:headline}, \ref{fig:serving} and~\ref{fig:xplat}).

Our observation is that many fast kernels are batch-invariant by themselves.
For those kernels the batch reaches a request's bits only through kernel selection, so we only need to fix the batch-dependent values that kernel selection reads.
Writing BI for batch invariance,
\begin{equation}
\underbrace{\text{kernel BI}}_{\text{user-supplied}} \;\wedge\; \underbrace{\text{selection BI}}_{\text{\sys}} \;\Longrightarrow\; \text{end-to-end BI},
\end{equation}
provided each request's output is mathematically a function of its own input (\S\ref{sec:overview}).
We present \sys. It keeps the fast kernels and makes kernel selection batch-invariant.
For each deployment and engine version, \sys decides which values in kernel selection to fix. We call them \emph{pins}.
For each workload, it picks the pins' values and a few deployment settings, because the fastest choice depends on the traffic.
These values form a \emph{commitment tuple}, fixed while serving.

Both decisions are hard to make by hand.
The source does not show which kernels agree bit for bit, and there are too many configurations to measure (\S\ref{sec:background}).
\sys automates both decisions and integrates into existing engines such as vLLM and SGLang (Figure~\ref{fig:overview}, \S\ref{sec:impl}).
We write kernel selection in \selir (Selection IR), an executable DSL.
The engine runs the user's selection program as its selector (\S\ref{sec:selir}).
A solver finds the fewest values in that program to pin so that no batch-dependent input reaches the selection.
It also finds the values each pin may legally take (\S\ref{sec:solver}).
A tuner picks the commitment tuple for the workload (\S\ref{sec:tuner}).
The user supplies batch-invariant kernels.

On multi-GPU deployments of Llama-3.1-70B and Qwen3-30B-A3B in vLLM and SGLang, \sys is batch-invariant as long as the commitment tuple stays the same.
Tuned per workload, \sys serves within 12\% of Stock's p50 latency at the geometric mean and is 2.4--5.7\tx faster than the engines' deterministic modes (\S\ref{sec:eval-perf}).
In SGLang it is as accurate on GSM8K as \sgdet (\S\ref{sec:eval-acc}).
On multiple types of workloads, one tuple for all loads of a type costs 1.1\% (Llama-3.1-70B) and 2.7\% (Qwen3-30B-A3B) more.
The tuple of another type can take 4.17\tx as long.

With seeded sampling committed, a trainer on the serving kernels and tuple matches every rollout logprob bit for bit on multiple models.
Rollout takes 1.03\tx Stock's time per token, and a whole RL step up to 1.18\tx.
Seeded rollouts of Qwen3-30B-A3B and Llama-3.1-70B on four GPUs also replay bit for bit under other batches and cache splits, and they match their rescoring.
On the BF16 recipe of~\citet{qi2025fp16}, Stock's GRPO~\citep{shao2024deepseekmath} run collapses while \sys's run stays stable (\S\ref{sec:eval-rl}).

We make the following contributions:
\begin{enumerate}[leftmargin=*]%
\item We give a sufficient condition for end-to-end batch invariance. With batch-invariant kernels, any \selir selection is batch-invariant on the deployment and engine version solved for once the values that separate it from the batch are pinned to legal values (\S\ref{sec:overview}, \S\ref{sec:solver}).
\item We design \selir, a selection language with a JIT cache. The solver analyzes its programs and the engine runs them (\S\ref{sec:selir}).
\item We build a solver that finds the fewest such pins and their legal values. A tuner commits a tuple per workload in minutes, and the tuple serves at near-stock speed (\S\ref{sec:solver}, \S\ref{sec:tuner}).
\item We implement \sys in vLLM and SGLang. We evaluate it across both engines, two GPU generations, dense and MoE models, and tensor-parallel (TP) degrees (\S\ref{sec:impl}, \S\ref{sec:eval}).
\item We show that a trainer integrated with \sys achieves on-policy RL training and matches RL rollouts bit for bit at near-stock speed (\S\ref{sec:eval-rl}).
\end{enumerate}

\section{Background and Motivation}
\label{sec:background}

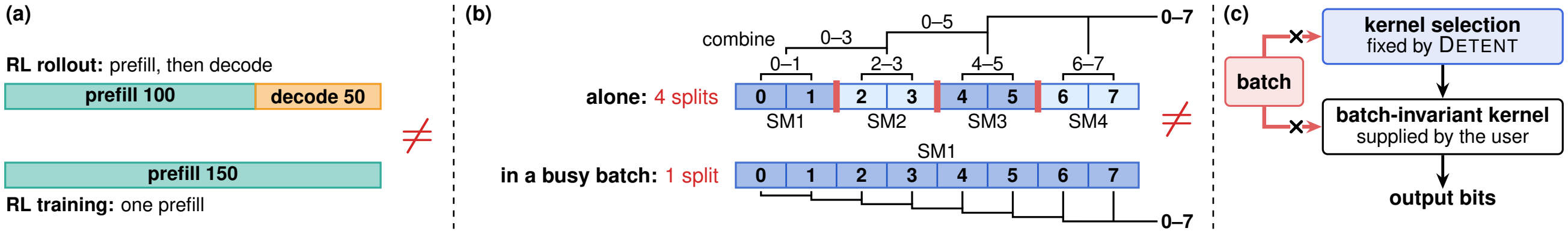
\begin{figure*}[t]
\definecolor{fwRed}{HTML}{D62828}
\definecolor{fwBlue}{HTML}{1D4ED8}
\definecolor{fwSlate}{HTML}{4A5D7E}
\definecolor{fwInk}{HTML}{000000}
\definecolor{fwPre}{HTML}{0F9D8A}
\definecolor{fwDec}{HTML}{F08C00}
\definecolor{fwGa}{HTML}{2459C9}
\definecolor{fwGb}{HTML}{5AA8F2}
\colorlet{fwRedS}{fwRed!75}
\colorlet{fwPreM}{fwPre!40}\colorlet{fwPreS}{fwPre!85}
\colorlet{fwDecM}{fwDec!40}\colorlet{fwDecS}{fwDec!85}
\colorlet{fwGaM}{fwGa!40}
\colorlet{fwGbF}{fwGb!20}
\colorlet{fwGaS}{fwGa!85}
\centering
\resizebox{\textwidth}{!}{%
\begin{tikzpicture}[
  font=\sffamily\fontsize{6.43}{7.37}\selectfont,
  ptitle/.style={font=\sffamily\fontsize{6.48}{7.56}\selectfont\bfseries, anchor=north west, inner sep=0pt, text=fwInk},
  lab/.style={font=\sffamily\fontsize{6.43}{7.37}\selectfont\bfseries, inner sep=0pt, text=fwInk},
  note/.style={font=\sffamily\fontsize{6}{7.03}\selectfont, inner sep=0pt, text=fwInk},
  inw/.style={font=\sffamily\fontsize{6}{6.86}\selectfont\bfseries, inner sep=0pt, text=fwInk},
  rsep/.style={draw=fwRedS, line width=2pt},
  brk/.style={draw=fwInk, line width=0.6pt},
  sep/.style={draw=fwInk, line width=0.5pt, dash pattern=on 2pt off 2pt},
  neq/.style={text=fwRed, font=\fontsize{12}{12}\selectfont\bfseries, inner sep=0pt},
  >={Stealth[length=1.7mm,width=1.4mm]},
]
\node[ptitle] at (0,-0.2486) {(a)};
\fill[fwPreM] (0,-1.3886) rectangle (2.7333,-1.1143);
\fill[fwDecM] (2.7333,-1.3886) rectangle (4.1,-1.1143);
\draw[overlay, fwPreS, line width=0.6pt] (0,-1.3886) rectangle (2.7333,-1.1143);
\draw[overlay, fwDecS, line width=0.6pt] (2.7333,-1.3886) rectangle (4.1,-1.1143);
\node[inw] at (1.3667,-1.2514) {prefill 100};
\node[inw] at (3.4167,-1.2514) {decode 50};
\fill[fwPreM] (0,-2.25) rectangle (4.1,-1.9757);
\draw[overlay, fwPreS, line width=0.6pt] (0,-2.25) rectangle (4.1,-1.9757);
\node[inw] at (2.05,-2.1129) {prefill 150};
\node[neq] at (4.5,-1.6821) {$\boldsymbol{\neq}$};%
\node[note, anchor=base west] at (0,-0.9771) {\textbf{RL rollout:} prefill, then decode};
\node[note, anchor=base west] at (0,-2.5071) {\textbf{RL training:} one prefill};
\draw[sep] (4.898,-0.2486) -- (4.898,-2.7);
\begin{scope}[xshift=5.019cm]
\node[ptitle] at (0,-0.2486) {(b)};
\foreach \i in {0,...,7} {
  \pgfmathtruncatemacro{\g}{mod(floor(\i/2),2)}
  \ifnum\g=0 \fill[fwGaM] ({2.96+0.55*\i},-1.3886) rectangle ({2.96+0.55*\i+0.55},-1.1143);
  \else      \fill[fwGbF] ({2.96+0.55*\i},-1.3886) rectangle ({2.96+0.55*\i+0.55},-1.1143); \fi
  \fill[fwGaM] ({2.96+0.55*\i},-2.25) rectangle ({2.96+0.55*\i+0.55},-1.9757);
}
\foreach \i in {1,...,7} {
  \draw[fwGa, line width=0.5pt] ({2.96+0.55*\i},-1.3886) -- ({2.96+0.55*\i},-1.1143);
  \draw[fwGa, line width=0.5pt] ({2.96+0.55*\i},-2.25) -- ({2.96+0.55*\i},-1.9757);
}
\draw[fwGaS, line width=0.6pt] (2.96,-1.3886) rectangle (7.36,-1.1143);
\draw[fwGaS, line width=0.6pt] (2.96,-2.25) rectangle (7.36,-1.9757);
\foreach \i in {0,...,7} {
  \node[inw] at ({3.235+0.55*\i},-1.2514) {\i};
  \node[inw] at ({3.235+0.55*\i},-2.1129) {\i};
}
\foreach \s in {4.06,5.16,6.26} { \draw[rsep] (\s,-1.4271) -- (\s,-1.0757); }
\foreach \k in {1,...,4} { \node[note, anchor=base] at ({2.96+0.55*(2*\k-1)},-1.5954) {SM\k}; }
\foreach \j in {0,...,3} {
  \pgfmathsetmacro{\xa}{3.235+1.1*\j}\pgfmathsetmacro{\xb}{\xa+0.55}
  \pgfmathtruncatemacro{\lo}{2*\j}\pgfmathtruncatemacro{\hi}{2*\j+1}
  \draw[brk] (\xa,-1.0757) -- (\xa,-1.0071) -- (\xb,-1.0071) -- (\xb,-1.0757);
  \node[note, anchor=south] at ({\xa+0.275},-0.9729) {\lo--\hi};
}
\draw[brk] (3.51,-0.79) -- (3.51,-0.7215) -- (4.61,-0.7215) -- (4.61,-0.55) -- (5.71,-0.55)
  -- (5.71,-0.3786) -- (7.566,-0.3786);
\draw[brk] (4.61,-0.79) -- (4.61,-0.7215);
\draw[brk] (5.71,-0.79) -- (5.71,-0.55);
\draw[brk] (6.81,-0.79) -- (6.81,-0.3786);
\node[note, anchor=south] at (4.06,-0.6872) {0--3};
\node[note, anchor=south] at (5.16,-0.5157) {0--5};
\node[inw, anchor=west] at (7.600,-0.3786) {0--7};
\node[note, anchor=base east] at (3.40,-0.6957) {combine};
\draw[brk] (3.235,-2.2886) -- (3.235,-2.3357)
  \foreach \k in {1,...,6} { -- ({3.235+0.55*\k},{-2.2886-0.04714*\k}) -- ({3.235+0.55*\k},{-2.2886-0.04714*(\k+1)}) }
  -- (7.085,-2.6186) -- (7.566,-2.6186);
\foreach \k in {1,...,7} { \draw[brk] ({3.235+0.55*\k},-2.2886) -- ({3.235+0.55*\k},{-2.2886-0.04714*\k}); }
\node[inw, anchor=west] at (7.600,-2.6186) {0--7};
\node[lab, anchor=base east] at (2.78,-1.3329) {alone: {\mdseries\color{fwRed}4 splits}};
\node[lab, anchor=base east] at (2.78,-2.1943) {in a busy batch: {\mdseries\color{fwRed}1 split}};
\node[neq] at (7.776,-1.5) {$\boldsymbol{\neq}$};%
\node[note, anchor=base] at (5.16,-1.926) {SM1};%
\end{scope}
\draw[sep] (13.193,-0.2486) -- (13.193,-2.7);
\begin{scope}[xshift=13.3cm]
\node[ptitle] at (0,-0.2486) {(c)};
\tikzset{fbox/.style={draw=#1!85, fill=#1!12, text=fwInk, rounded corners=1.5pt, line width=0.6pt,
  minimum height=6mm, inner sep=2pt, align=center, font=\sffamily\fontsize{6.43}{7.37}\selectfont\bfseries}}
\node[fbox=fwRed, draw=fwRedS, minimum width=0.85cm, minimum height=5.31mm] (bt) at (0.45,-1.0893) {batch};
\node[fbox=fwBlue, minimum width=2.62cm] (sl) at (2.4,-0.5957) {kernel selection\\[-0.86pt]{\fontsize{6}{6.86}\selectfont\mdseries fixed by {\fontsize{6.86}{7.71}\selectfont\sys}}};
\node[fbox=fwInk, draw=fwInk, fill=white, minimum width=2.62cm] (kn) at (2.4,-1.5831) {batch-invariant kernel\\[-0.86pt]{\fontsize{6}{6.86}\selectfont\mdseries supplied by the user}};%
\node[lab] (bits) at (2.4,-2.3766) {output bits};
\draw[fwRedS, line width=1.2pt] (bt.north) |- (0.735,-0.5957);
\draw[->, fwRedS, line width=1.2pt] (0.865,-0.5957) -- (sl.west);
\draw[fwRedS, line width=1.2pt] (bt.south) |- (0.735,-1.5831);
\draw[->, fwRedS, line width=1.2pt] (0.865,-1.5831) -- (kn.west);
\foreach \y in {-0.5957,-1.5831} {
  \draw[fwInk, line width=0.9pt, line cap=round] (0.8,\y) ++(-0.0575,-0.0575) -- ++(0.115,0.115)
                                                 (0.8,\y) ++(-0.0575,0.0575) -- ++(0.115,-0.115);}
\draw[->, fwInk, line width=0.9pt] (sl) -- (kn);
\draw[->, fwInk, line width=0.9pt] (kn) -- (bits);
\end{scope}
\end{tikzpicture}}%
\caption{The batch regroups a request's additions, so its bits change. (a)~Where prefill ends. (b)~Kernel selection sets the split count; each split sums its KV blocks, and a combine adds the sums in order. (c)~\sys fixes selection; the user supplies kernels.}
\label{fig:why}
\end{figure*}

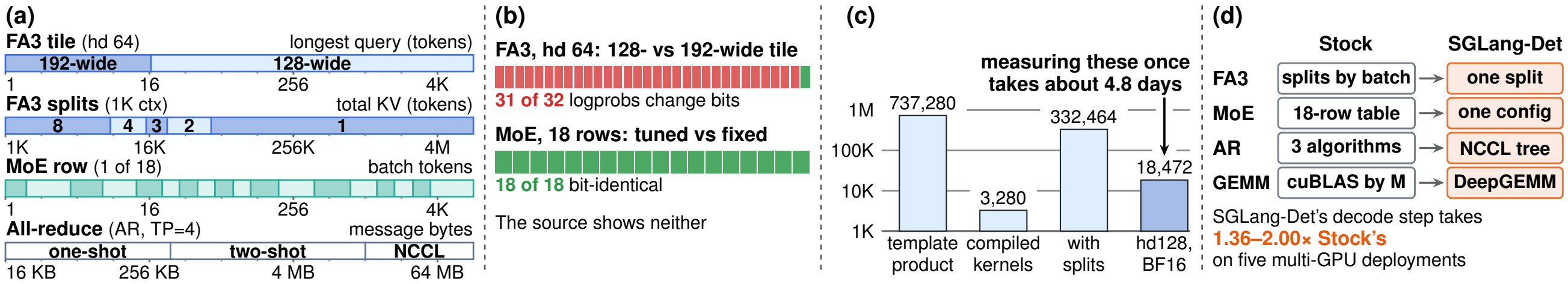
\begin{figure*}[t]
\definecolor{fhRedO}{HTML}{D62828}
\definecolor{fhGreenO}{HTML}{2B9A48}
\definecolor{fhInk}{HTML}{1F2328}
\colorlet{fhText}{black}
\definecolor{fhAO}{HTML}{2459C9}
\definecolor{fhBO}{HTML}{5AA8F2}
\definecolor{fhTealO}{HTML}{0F9D8A}
\definecolor{fhTealLO}{HTML}{5CC4B5}
\definecolor{fhSaO}{HTML}{4A5D7E}
\definecolor{fhStockO}{HTML}{5F6B7A}
\definecolor{fhDetO}{HTML}{E8590C}
\colorlet{fhRed}{fhRedO!75}
\colorlet{fhRedD}{fhRedO}
\colorlet{fhGreen}{fhGreenO!85}
\colorlet{fhGreenD}{fhGreenO}
\colorlet{fhA}{fhAO!40}
\colorlet{fhB}{fhBO!20}
\colorlet{fhAS}{fhAO!85}
\colorlet{fhAD}{fhAO}
\colorlet{fhTeal}{fhTealO!40}
\colorlet{fhTealL}{fhTealLO!20}
\colorlet{fhTealS}{fhTealO!85}
\colorlet{fhSS}{fhSaO!85}
\colorlet{fhStock}{fhStockO!85}
\colorlet{fhStockF}{white}
\colorlet{fhStockD}{fhStockO}
\colorlet{fhDet}{fhDetO!85}
\colorlet{fhDetF}{fhDetO!12}
\colorlet{fhDetD}{fhDetO}
\centering
\resizebox{\textwidth}{!}{%
\begin{tikzpicture}[y=1.1cm,%
  >={Stealth[length=1.6mm,width=1.3mm]},
  font=\sffamily\fontsize{7.92}{9.09}\selectfont,
  ptitle/.style={font=\sffamily\fontsize{9.5}{11.09}\selectfont\bfseries, anchor=north west, inner sep=0pt, text=fhText},
  lab/.style={font=\sffamily\fontsize{7.92}{9.09}\selectfont\bfseries, inner sep=0pt, text=fhText},
  note/.style={font=\sffamily\fontsize{7.39}{8.67}\selectfont, inner sep=0pt, text=fhText},
  inw/.style={font=\sffamily\fontsize{7.39}{8.45}\selectfont\bfseries, inner sep=0pt, text=fhText},
  qty/.style={font=\sffamily\fontsize{7.39}{8.45}\selectfont, inner sep=0pt, text=fhText, anchor=base east},
  rname/.style={font=\sffamily\fontsize{7.92}{9.09}\selectfont\bfseries, inner sep=0pt, text=fhText, anchor=base west},
  tick/.style={font=\sffamily\fontsize{7.39}{8.45}\selectfont, inner sep=0pt, text=fhText, anchor=north},
  sep/.style={draw=fhInk!75, line width=0.65pt, dash pattern=on 2pt off 2pt},
]
\begin{scope}[xscale=1.4600]%
\node[ptitle] at (0,3.844) {(a)};
\def\bh{0.213}
\newcommand{\fhaxis}[5]{%
  \foreach \oc in {1,...,12} { \draw[fhInk!55, line width=0.65pt] ({4.5*\oc/13},#1) -- ({4.5*\oc/13},#1-0.027); }
  \foreach \oc in {0,4,8,12} { \draw[fhInk!70, line width=0.65pt] ({4.5*\oc/13},#1) -- ({4.5*\oc/13},#1-0.04); }
  \node[tick, anchor=north west] at (0,#1-0.071) {#2};
  \node[tick] at ({4.5*4/13},#1-0.071) {#3};
  \node[tick] at ({4.5*8/13},#1-0.071) {#4};
  \node[tick] at ({4.5*12/13},#1-0.071) {#5};
}
\def\yb{3.00}
\node[rname] at (0,\yb+\bh+0.089) {FA3 tile {\fontsize{7.39}{8.45}\selectfont\mdseries\color{fhText}(hd 64)}};
\node[qty] at (4.5,\yb+\bh+0.089) {longest query (tokens)};
\filldraw[fill=fhA, draw=fhAS, line width=0.65pt, overlay] (0,\yb) rectangle (1.4,\yb+\bh);   \node[inw] at (0.7,\yb+\bh/2) {192-wide};
\filldraw[fill=fhB, draw=fhAS, line width=0.65pt, overlay] (1.4,\yb) rectangle (4.5,\yb+\bh); \node[inw] at (2.95,\yb+\bh/2) {128-wide};
\fhaxis{\yb}{1}{16}{256}{4K}
\def\yb{2.20}
\node[rname] at (0,\yb+\bh+0.089) {FA3 splits {\fontsize{7.39}{8.45}\selectfont\mdseries\color{fhText}(1K ctx)}};
\node[qty] at (4.5,\yb+\bh+0.089) {total KV (tokens)};
\foreach \a/\b/\n/\c/\t in {0/1.01/8/fhA/fhText, 1.01/1.354/4/fhB/fhText, 1.354/1.556/3/fhA/fhText, 1.556/1.979/2/fhB/fhText, 1.979/4.5/1/fhA/fhText} {
  \filldraw[fill=\c, draw=fhAS, line width=0.65pt, overlay] (\a,\yb) rectangle (\b,\yb+\bh); \node[inw, text=\t] at ({(\a+\b)/2},\yb+\bh/2) {\n};
}
\fhaxis{\yb}{1K}{16K}{256K}{4M}
\def\yb{1.40}
\node[rname] at (0,\yb+\bh+0.089) {MoE row {\fontsize{7.39}{8.45}\selectfont\mdseries\color{fhText}(1 of 18)}};
\node[qty] at (4.5,\yb+\bh+0.089) {batch tokens};
\foreach \a/\b [count=\i] in {0/0.202, 0.202/0.626, 0.626/0.935, 0.935/1.261, 1.261/1.508, 1.508/1.673,
    1.673/1.848, 1.848/2.015, 2.015/2.191, 2.191/2.359, 2.359/2.627, 2.627/2.972, 2.972/3.318,
    3.318/3.573, 3.573/3.741, 3.741/3.919, 3.919/4.087, 4.087/4.5} {
  \ifodd\i \filldraw[fill=fhTeal, draw=fhTealS, line width=0.65pt, overlay] (\a,\yb) rectangle (\b,\yb+\bh); \else \filldraw[fill=fhTealL, draw=fhTealS, line width=0.65pt, overlay] (\a,\yb) rectangle (\b,\yb+\bh); \fi
}
\fhaxis{\yb}{1}{16}{256}{4K}
\def\yb{0.60}
\node[rname] at (0,\yb+\bh+0.089) {All-reduce {\fontsize{7.39}{8.45}\selectfont\mdseries\color{fhText}(AR, TP=4)}};
\node[qty] at (4.5,\yb+\bh+0.089) {message bytes};
\filldraw[fill=white, draw=fhSS, line width=0.65pt, overlay] (0,\yb) rectangle (1.587,\yb+\bh);     \node[inw] at (0.7935,\yb+\bh/2) {one-shot};
\filldraw[fill=white, draw=fhSS, line width=0.65pt, overlay] (1.587,\yb) rectangle (3.462,\yb+\bh); \node[inw] at (2.5245,\yb+\bh/2) {two-shot};
\filldraw[fill=white, draw=fhSS, line width=0.65pt, overlay] (3.462,\yb) rectangle (4.5,\yb+\bh);   \node[inw] at (3.981,\yb+\bh/2) {NCCL};
\fhaxis{\yb}{16 KB}{256 KB}{4 MB}{64 MB}
\end{scope}
\draw[sep] (6.730,0.467) -- (6.730,3.844);
\begin{scope}[xshift=16.940cm]%
\node[ptitle] at (0,3.844) {(d)};
\tikzset{sbox/.style={fill=#1F, draw=#1, line width=0.7pt, outer sep=0.2pt, text=fhText, rounded corners=1.5pt, minimum height=4.30mm, inner sep=1.5pt,
  align=center, font=\sffamily\fontsize{7.39}{8.45}\selectfont\bfseries}}
\node[lab, anchor=base] at (1.88,3.267) {Stock};
\node[lab, anchor=base east] at (4.93,3.267) {\sgdet};
\foreach \y/\f/\s/\d in {2.911/FA3/splits by batch/one split, 2.467/MoE/18-row table/one config, 2.022/AR/3 algorithms/NCCL tree, 1.578/{\fontsize{7.39}{8.45}\selectfont GEMM}/cuBLAS by M/DeepGEMM} {
  \node[lab, anchor=west] at (0,\y) {\f};
  \node[sbox=fhStock, minimum width=1.92cm, anchor=west] (s) at (0.92,\y) {\s};%
  \node[sbox=fhDet, minimum width=1.62cm, anchor=east] (d) at (4.93,\y) {\d};%
  \draw[->, fhInk!70, line width=0.8pt, shorten <=1.5pt, shorten >=1pt] (s.east) -- (d.west);
}
\node[note, anchor=north west, align=left] at (0,1.249) {\sgdet's decode step takes};
\node[lab, text=fhDetD, anchor=north west] at (0,0.973) {1.36--2.00\texttimes\ Stock's};
\node[note, anchor=north west] at (0,0.698) {on five multi-GPU deployments};
\end{scope}
\draw[sep] (11.450,0.467) -- (11.450,3.844);
\begin{scope}[xshift=11.800cm]%
\node[ptitle] at (0,3.844) {(c)};
\foreach \e/\y in {1K/0.96, 10K/1.476, 100K/1.991, 1M/2.507} {
  \node[note, anchor=east] at (0.40,\y) {\e};
}
\draw[fhInk!70, line width=0.65pt] (0.5,0.96) -- (4.83,0.96);
\foreach \x/\v/\h/\la/\lb/\c [count=\i] in {
  1.080/737{,}280/2.439/template/product/fhB,
  2.210/3{,}280/1.226/compiled/kernels/fhB,
  3.340/332{,}464/2.260/with/splits/fhB,
  4.470/18{,}472/1.613/hd128{,}/BF16/fhA} {
  \filldraw[fill=\c, draw=fhInk, line width=0.5pt] (\x-0.33,0.96) rectangle (\x+0.33,\h);%
  \node[note, anchor=south, fill=white, inner xsep=1pt, inner ysep=0.4pt] (v\i) at (\x,\h+0.027) {\v};
  \node[note, anchor=north, align=center] at (\x,0.907) {\la\\\lb};
}
\node[lab, anchor=south east, align=center] (m) at (4.84,2.693) {measuring these once\\takes about 4.8 days};%
\draw[->, black, line width=0.9pt] (4.470,2.667) -- (4.470,1.920);
\begin{pgfonlayer}{background}
\tikzset{fhgrid/.style={fhInk!55, line width=0.65pt}}
\draw[fhgrid] (0.5,0.96) -- (4.83,0.96);
\begin{scope}[even odd rule]%
  \clip (0.45,1.316) rectangle (4.90,1.636) ([shift={(-3pt,-3pt)}]v2.south west) rectangle ([shift={(3pt,3pt)}]v2.north east)
    (4.80,1.404) rectangle (4.90,1.547);
  \draw[fhgrid] (0.5,1.476) -- (4.83,1.476);
\end{scope}
\begin{scope}[even odd rule]%
  \clip (0.45,1.831) rectangle (4.90,2.151) ([shift={(-0.65mm-3pt,-3pt)}]4.470,1.920) rectangle ([shift={(0.65mm+3pt,3pt)}]4.470,2.080);
  \draw[fhgrid] (0.5,1.991) -- (4.83,1.991);
\end{scope}
\begin{scope}[even odd rule]%
  \clip (0.45,2.347) rectangle (4.90,2.667) ([shift={(-3pt,-3pt)}]v1.south west) rectangle ([shift={(3pt,3pt)}]v1.north east)
    ([shift={(-3pt,-3pt)}]v3.south west) rectangle ([shift={(3pt,3pt)}]v3.north east)
    ([xshift=-0.45pt-3pt]4.470,2.382) rectangle ([xshift=0.45pt+3pt]4.470,2.631);
  \draw[fhgrid] (0.5,2.507) -- (4.83,2.507);
\end{scope}
\end{pgfonlayer}
\end{scope}
\draw[sep] (16.800,0.467) -- (16.800,3.844);
\begin{scope}[xshift=6.870cm]%
\node[ptitle] at (0,3.844) {(b)};
\def\bw{4.420}
\node[lab, anchor=south west] at (0,3.142) {FA3, hd 64: 128- vs 192-wide tile};
\foreach \i in {0,...,31} {
  \pgfmathsetmacro{\xl}{\i==0 ? 0 : \bw*\i/32+0.01147}
  \pgfmathsetmacro{\xr}{\i==31 ? \bw : \bw*(\i+1)/32-0.01147}
  \ifnum\i<31 \fill[fhRed] (\xl,2.787) rectangle (\xr,3.071);
  \else       \fill[fhGreen] (\xl,2.787) rectangle (\xr,3.071); \fi
}
\node[note, anchor=north west] at (0,2.733) {\textcolor{fhRedD}{\textbf{31 of 32}} logprobs change bits};
\node[lab, anchor=south west] at (0,2.058) {MoE, 18 rows: tuned vs fixed};
\foreach \i in {0,...,17} {
  \pgfmathsetmacro{\xl}{\i==0 ? 0 : \bw*\i/18+0.01147}
  \pgfmathsetmacro{\xr}{\i==17 ? \bw : \bw*(\i+1)/18-0.01147}
  \fill[fhGreen] (\xl,1.702) rectangle (\xr,1.987);
}
\node[note, anchor=north west, align=left] at (0,1.649) {\textcolor{fhGreenD}{\textbf{18 of 18}} bit-identical};
\node[note, anchor=north west] at (0,1.169) {The source shows neither};%
\end{scope}
\end{tikzpicture}}\par
\caption{Kernel selection reads the batch, and only measurement shows which kernels agree. (a)~Where four choices switch (log axes; H100; tile on Llama-3.2-1B, MoE row on Qwen3-30B-A3B, splits and all-reduce on Llama-3.1-70B under CUDA graphs). (b)~Agreement is measured. (c)~Too many FA3 configurations on H100 to measure (log scale). (d)~Fixing choices is slow.}
\label{fig:hardness}
\end{figure*}

\paragraph{Observation 1: selection reads the batch.}
Floating-point addition is not associative, so a kernel's bits depend on its reduction order.
A request's output is \emph{batch-invariant} when its bits do not depend on the batch it runs in~\citep{thinkingmachines2025}.
The batch gets in by several routes (Figure~\ref{fig:why}).
Kernel selection picks each kernel and its parameters.
Each pick fixes how a request's values are grouped and added.
FlashAttention-3 (FA3; \citealp{shah2024fa3}) sets its split count (Figure~\ref{fig:why}b) and tile width (Figure~\ref{fig:hardness}a) from the batch.
The fused MoE takes the row of its tuned configuration table nearest the batch's token count, and the all-reduce picks an algorithm by the bytes the batch sends (Figure~\ref{fig:hardness}a).
Across steps, the prefix cache and the scheduler divide a request's tokens between prefill and decode.
If a request prefills 100 tokens and then decodes 50, its keys are grouped differently than if it prefilled all 150 (Figure~\ref{fig:why}a).
An RL rollout prefills and then decodes, while training prefills everything.
Batch-invariant kernels fix the reduction order once these choices are made, but not the choices themselves (Figure~\ref{fig:why}c).

\paragraph{Observation 2: agreement must be measured.}
\sgdet and \vldet fix these choices.
\sgdet's decode step then takes 1.36--2.00\tx Stock's on five multi-GPU deployments (Figure~\ref{fig:hardness}d).
Keeping faster kernels instead requires knowing which of them agree bit for bit.
Only measurement shows that (Figure~\ref{fig:hardness}b).
At head dimension 64, two tile widths (Figure~\ref{fig:hardness}b) disagree on 31 of 32 logprobs.
On Qwen3-30B-A3B, by contrast, the 18 tuned MoE configurations agree bit for bit with the one \sgdet hard-codes.
Nothing in the source tells these two cases apart, and the answer changes with the deployment.

\paragraph{Observation 3: too many configurations to measure.}
Measuring each of FA3's 18,472 configurations once would take about 4.8 days of a four-GPU node (Figure~\ref{fig:hardness}c; Appendix~\ref{app:space}).
\sys fixes what kernel selection reads and never relies on two kernels agreeing (Observations 1 and 2).
It finds what to fix by analyzing the selection program (Observation 3).

\section{Overview}
\label{sec:overview}

\begin{figure*}[t]
\centering
\includegraphics{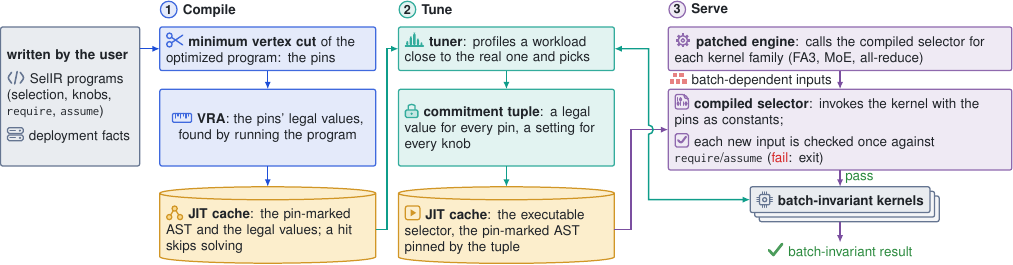}%
\caption{\sys isolates kernel selection from the batch.}
\label{fig:overview}
\vspace{-8pt}%
\end{figure*}

\paragraph{The condition.}
A request's output is
\begin{equation}
y = F(x, s;\ M, D, T;\ B),
\end{equation}
with prompt $x$, sampling seed $s$, model code and weights $M$, deployment $D$ (engine and its build, GPU, parallelism, \ldots), commitment tuple $T$ and batch state $B$ (co-running requests, row, prefix-cache hit, \ldots).
We call it batch-invariant when $F(x, s; M, D, T; B) = F(x, s; M, D, T; B')$ for all $B, B'$. CoRun states a similar contract~\citep{corun2026}.

\emph{Kernel batch invariance}: each candidate kernel, given its parameters, reduces a request's values in an order that depends only on that request.
\emph{Selection batch invariance}: the values a kernel selection returns, its \emph{selection points}, depend only on $M$, $D$, $T$ and the request.
Like prior work, we assume each request's output is mathematically a function of its own input.

A \emph{kernel family} is the set of kernels one \selir program selects among.
A kernel with its parameters fixed, such as one FA3 tile width and split count, counts as one kernel.
For each family, the solver returns \emph{pins}.
A pin is a value of the program that \sys replaces with a constant, such as FA3's split count.
An \emph{assignment} gives every pin a value.
It is \emph{legal} if every precondition holds on the inputs the solver judges.
\phantomsection\label{sec:overview-knob}A \emph{knob} is a deployment setting that no input determines, such as the all-reduce workspace (push or pull) or the padding of the matrix multiplication (GEMM).
The tuner then picks a \emph{commitment tuple}: a legal value for every pin and a setting for every knob.
For a \emph{workload}, the commitment tuple is fixed before deployment and does not change while serving, so the batch's composition never affects kernel selection.

\paragraph{The pipeline.}
The solver finds the pins and their legal values in a SelIR program once per deployment and engine version (\S\ref{sec:solver}).
The tuner picks a commitment tuple per workload (\S\ref{sec:tuner}; Figure~\ref{fig:overview}).
During serving, the same \selir program selects kernels with the tuple's values pinned. The pins block every path from the batch to a selection point (\S\ref{sec:solver}), so selection batch invariance holds by construction.
Kernel batch invariance is the user's responsibility (\S\ref{sec:kernels}).
\phantomsection\label{sec:overview-rl}%
An RL trainer that runs the same kernels under the same tuple differs from the server only in the batch, so it matches the rollout's logprobs. Admitting only seeded sampling also fixes the sampled tokens (\S\ref{sec:eval-rl}).

\section{\selir}
\label{sec:selir}

Each kernel call site uses a \selir program in place of the engine's selection code.
Because the solver analyzes this same program, pinning a value in \selir also pins it in the running engine.

\paragraph{The language.}
\selir is a Python-subset DSL with Python's semantics.
Its programs are pure functions over records and scalars, with conditionals and reductions over static collections.
Batch-dependent inputs are marked \code{Runtime}, or \code{Unpinnable} if they cannot be pinned.
Each value the program returns is a \emph{selection point}.
\code{knob} declares a deployment setting (\S\ref{sec:overview-knob}).
\code{require} and \code{assume} together state when the program's selection is valid.
The user never names pins or their candidate values.
A program may transcribe an engine's selection or state a new policy for the user's kernels.
Our programs transcribe an engine's selection, then adapt its heuristics to the kernels \sys selects among.
For example, the FA3 program adds a split stride (\S\ref{sec:kernels}).
The engine's heuristic makes no such choice.
Both engines we evaluate run these programs.
Selections for other kernels and engines are written the same way.

\paragraph{Optimization and lowering.}
Most selection inputs are known before serving: the model's head dimension and head counts, the data type, the GPU architecture, the degree of parallelism, the page size, the maximum context, etc.
A few change at run time, such as the batch's token count and KV lengths.
We therefore partially evaluate each program for its deployment~\citep{jones1993pe}.
Compiler optimizations lower it to a single dependence graph.
Its edges capture every dependence on the batch, including control dependences.
The graph's values are simplified as far as the deployment facts allow.
With constant folding alone, FA3's cut has 12 pins instead of 3 (\S\ref{sec:eval-solver}).

\subsection{Compiling and Caching Selections}

As with TorchDynamo's code cache~\citep{ansel2024pytorch2}, TVM's tuning records~\citep{chen2018tvm} and TensorRT's engines~\citep{tensorrt2026}, we compile a selection once and reuse it (Figure~\ref{fig:overview}).
A solve is stored on disk under a hash of everything that can change its answer.
A hit in this JIT cache skips solving (\S\ref{sec:solver}).
For FA3, a hit takes about 20\,ms and a cold solve of one committed tuple about 5\,s.
The compiled selector has the pins as constants.
While serving, each new combination of \code{Runtime} inputs is checked once against every \code{require} and \code{assume}.
The compiled selector runs at every call.
Unlike a failed shape guard in \code{torch.compile}, a new batch never triggers a recompile.

\section{Solver}
\label{sec:solver}

\begin{figure}[t]
\centering
\includegraphics[width=\columnwidth]{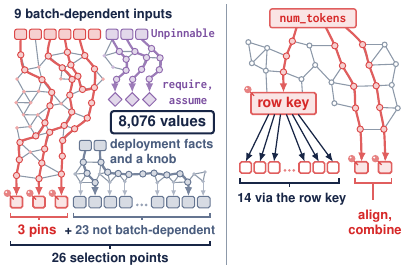}%
\caption{The solver separates kernel selection from the batch with a few pins. Left, the whole selection of FA3 on Llama-3.1-70B; right, the MoE's on Qwen3-30B-A3B. In FA3, red inputs are \code{Runtime} or computed at launch, purple ones other \code{Unpinnable} inputs, each value is tinted by the inputs it derives from, and gray values derive only from deployment facts and a knob.}
\label{fig:solver}
\end{figure}

The solver decides which values of a \selir program to pin and which values each pin may take. It runs once per deployment and engine version.

\paragraph{The cut.}
After optimization, a selection is a dependence graph from its inputs to its selection points.
For FA3 on Llama-3.1-70B, the graph has 8,076 values and 26 selection points, 3 of them batch-dependent (Figure~\ref{fig:solver}).
The batch must not reach any selection point.
Pinning a value turns it into a constant, which no path crosses.
The solver therefore pins a vertex cut, a set of values that blocks every path from a batch-dependent input to a selection point.
Pinning every batch-dependent selection point always gives such a cut, but each pin adds a dimension to the tuner's search space.
The solver instead takes a weighted minimum cut via maximum flow~\citep{edmonds1972}.
The weights favor the fewest pins and, among cuts of equal size, pins near the selection points.
On FA3 the cut falls on batch-dependent selection points.
On the MoE it falls inside the program, on the row key (the token count of the tuned configuration row it uses).
This pin fixes 14 of the MoE's 16 batch-dependent selection points.

\paragraph{Legal values.}
\emph{Pin Value Range Analysis} (VRA) finds the values each pin may take.
A pin's candidates are the values it can take in the program.
VRA runs the program with an assignment of candidates substituted.
It judges the assignment legal if every \code{require} and \code{assume} holds at every input point it tries.
It tries all points when the input domains are small and a fixed sample otherwise.
While serving, the selector checks these conditions on each new input.
If one fails, the engine stops with an error and runs no other kernel.

\section{Tuner}
\label{sec:tuner}

The fastest tuple depends on the workload, for example on its prefix-hit rate, concurrency and prompt lengths (\S\ref{sec:eval-perf}).
The tuner picks a commitment tuple for each workload.

\begin{figure}[t]
\centering
\definecolor{cFA}{HTML}{0B7285}\definecolor{cMoE}{HTML}{2F9E44}\definecolor{cAR}{HTML}{9C36B5}\definecolor{cGM}{HTML}{495057}
\definecolor{tBlue}{HTML}{1D4ED8}
\colorlet{cFAF}{cFA!12}\colorlet{cFAS}{cFA!85}
\colorlet{cMoEF}{cMoE!14}\colorlet{cMoES}{cMoE!85}
\colorlet{cARF}{cAR!12}\colorlet{cARS}{cAR!85}
\colorlet{cGMF}{white}\colorlet{cGMS}{cGM!85}
\newcommand{\famtag}[2]{\tikz[baseline=(t.base)]\node[fill=#1F, draw=#1S, line width=0.6pt, rounded corners=1.5pt,
  inner xsep=2pt, inner ysep=1.2pt, minimum width=25pt, text=black, font=\sffamily\bfseries\fontsize{7}{8}\selectfont] (t) {#2};}
{\sffamily\fontsize{7}{8.4}\selectfont\renewcommand{\arraystretch}{1.28}\setlength{\tabcolsep}{6pt}%
\begin{tabular}{@{}l r r r@{}}
\toprule
\textbf{family} & \textbf{candidates} & \textbf{legal} & \textbf{measured}\\
\midrule
\famtag{cFA}{FA3}  & 7,168 & 2,976 & 48\\
\famtag{cMoE}{MoE} & 76    & 36    & 20\\
\famtag{cAR}{AR}   & 204   & 3     & 3\\
\famtag{cGM}{GEMM} & 5     & 5     & 5\\
\multicolumn{4}{@{}l}{\ldots\ any other family written in \selir}\\
\midrule
one family at a time (sum)     & & \textbf{\textcolor{tBlue}{3,020}} & \textbf{\textcolor{tBlue}{76}}\\
all families jointly (product) & & 1,607,040 & \\
\bottomrule
\end{tabular}}
\caption{The tuner searches one kernel family at a time. A case on Qwen3-30B-A3B: candidates multiply a family's pin and knob values, VRA keeps the legal ones, and the tuner measured 76 of the 3,020 legal assignments.}
\label{fig:tuner}
\end{figure}

\paragraph{Profile.}
For each workload, we tune a commitment tuple once using a profiling workload with similar arrivals.
The target workload then runs with the fixed tuple.
Tuning first serves the profiling workload with the \emph{default tuple}.
It gives each pin the legal value the unpinned program takes most often and each knob its default.
For each kernel family $f$, the tuner records in $W_f$ the family's distinct calls $c$ with their batch-dependent quantities and counts $n$.

\paragraph{Additivity.}
A tuple $a = (a_f)_f$ gives each family $f$ a legal assignment $a_f \in L_f$.
Since legality is judged per family, the legal tuples are $L = \prod_f L_f$.
We make three assumptions.
(i) Kernels of different families run back to back on one stream without overlap.
(ii) Because the recorded workload fixes batching and outputs, a tuple changes only which kernels each family selects.
(iii) The time $t_f(c, a_f)$ of a call $c$ depends only on the call and its family's assignment.
Then, with $C$ the time of all other work, the workload's time under $a$ is
\begin{equation}
T(a) = C + \sum_f T_f(a_f), \quad T_f(a_f) = \!\!\sum_{(c,n) \in W_f}\!\! n \cdot t_f(c, a_f).
\end{equation}
Hence
\begin{equation*}
\arg\min_{a \in L} T(a) = \bigl(\arg\min_{a_f \in L_f} T_f(a_f)\bigr)_f.
\end{equation*}
Proof: the per-family minimizers $a^\ast_f$ form a tuple $a^\ast \in L$.
Every $a \in L$ has $T(a) \ge C + \sum_f T_f(a^\ast_f) = T(a^\ast)$.
The tuner therefore searches $\sum_f |L_f|$ assignments instead of $\prod_f |L_f|$.
On Qwen3-30B-A3B, it searches 3,020 instead of 1,607,040 (Figure~\ref{fig:tuner}).
If the sum holds only within a relative error $\delta$, i.e.\ $|T(a) - \hat{T}(a)| \le \delta\,T(a)$ for every legal $a$ with $\hat{T}(a) = C + \sum_f T_f(a_f)$, then for every legal $a$
\begin{equation*}
T(a^\ast) \le \tfrac{\hat{T}(a^\ast)}{1-\delta} \le \tfrac{\hat{T}(a)}{1-\delta} \le \tfrac{1+\delta}{1-\delta}\,T(a).
\end{equation*}
On Qwen3-30B-A3B, the largest interaction between families was $\delta = 1.04\%$ of the measured GPU time per output token.
The per-family choice is then within 2.1\% of the best tuple.
The assumptions fail if an engine overlaps two families (e.g., kernel fusions) or if faster kernels change how arriving requests are batched.
The user must therefore declare kernels that overlap as one family.

\paragraph{Search.}
The tuner searches each family's legal grid and measures only a small part of it.
The grid's axes are the family's pins and the knobs that only its selection reads.
Each candidate is a repin that swaps the pins' constants and regenerates the selector without lowering or solving again.
On Llama-3.1-70B's exhaustive grids, which were not used to design the tuner, the tuner measures 8.0\% of the candidates and ends 0.04\% from the optimum on average (0.75\% at worst).

\section{Implementation}
\label{sec:impl}

\sys's solver with VRA, selector runtime, tuner and engine integrations are implemented in Python.
So are the \selir programs for FA3, fused MoE, all-reduce and GEMM.
The all-reduce kernels and FA3 modifications are implemented in CUDA.
In each engine, a patch routes kernel selection at the FA3, MoE and all-reduce call sites through the selector.
The rest of \sys runs unchanged.
We run SGLang (a 2026 development version) and vLLM 0.26.0.
For RL, a second patch outside \selir commits seeded sampling.
An RL trainer uses the serving kernels.

\paragraph{Kernels.}
\phantomsection\label{sec:kernels}
\sys uses the same kernels in both engines: DeepGEMM~\citep{deepgemm2025} for matrix multiplication, Triton fused-MoE kernels, our all-reduce kernels and a rebuilt FA3.
The rebuilt FA3 keeps split-KV and follows the fixed split-size principle~\citep{thinkingmachines2025}, so a request's split boundaries do not depend on the batch.
Our all-reduce kernels sum each element in a fixed rank order.
The tuner also picks two knobs like pins: FA3's split stride and the GEMM padding.

\section{Evaluation}
\label{sec:eval}

We ask what solving and tuning cost (\S\ref{sec:eval-kernels}), how close \sys's batch-invariant serving comes to Stock in vLLM and SGLang (\S\ref{sec:eval-perf}), whether it keeps accuracy (\S\ref{sec:eval-acc}), how it compares with LLM-42 (\S\ref{sec:eval-llm42}), whether a trainer matches its rollouts (\S\ref{sec:eval-rl}) and how far the solver carries to other deployments (\S\ref{sec:eval-port}).
We ablate the solver's optimizations (\S\ref{sec:eval-kernels}), the tuner (\S\ref{sec:eval-perf}, \S\ref{sec:eval-port}) and our kernels (Appendix~\ref{app:ablation}).

\subsection{Setup}
\label{sec:eval-setup}

\begin{table}[t]
\caption{Arms (\S\ref{sec:eval-setup}). BI: batch-invariant. Stock: each engine as shipped, deterministic mode off. \sgdet, \vldet: the engines' deterministic modes (Appendix~\ref{app:det}). MoE: Qwen3-30B-A3B only. No arm uses FlashInfer's all-reduce fusion; Appendix~\ref{app:ablation} ablates it and the matrix-multiplication and all-reduce kernels.}
\label{tab:arms}
\vskip 0.15in%
\centering
\scriptsize
\newcommand{\ac}[2]{\parbox[t]{#1}{\raggedright\strut #2\strut}}
\setlength{\tabcolsep}{2.4pt}
\begin{tabular}{@{}lclllll@{}}
\toprule
Arm & BI & \ac{0.42in}{Kernel selection} & \ac{0.5in}{Matrix mult.} & \ac{0.42in}{Attention split} & \ac{0.45in}{All-reduce} & \ac{0.4in}{MoE row key} \\
\midrule
Stock & \no & \ac{0.42in}{stock heuristics} & \ac{0.5in}{cuBLAS} & \ac{0.42in}{by the batch} & \ac{0.45in}{custom, by the batch's bytes} & \ac{0.4in}{nearest the batch} \\\addlinespace[2pt]
\ac{0.56in}{\sgdet\\ \vldet} & \ok & \ac{0.42in}{one kernel per operator} & \ac{0.5in}{DeepGEMM\\(Triton in vLLM)} & \ac{0.42in}{none} & \ac{0.45in}{NCCL tree} & \ac{0.4in}{fixed} \\\addlinespace[2pt]
\sys & \ok & \ac{0.42in}{\selir, pinned} & \ac{0.5in}{DeepGEMM (tuned pad)} & \ac{0.42in}{fixed stride or none} & \ac{0.45in}{ours, pinned} & \ac{0.4in}{pinned} \\
\bottomrule
\end{tabular}
\end{table}

\paragraph{Deployments and arms.}
The main deployments run Llama-3.1-70B-Instruct~\citep{dubey2024llama3} and Qwen3-30B-A3B (MoE; \citealp{yang2025qwen3}).
Both have head dimension 128 and are served by vLLM and SGLang in bf16 with tensor parallelism (TP) over four H100 GPUs.
\sys runs unchanged in both engines.
Only the deployment facts in its specs change (in vLLM, 16-token pages, its CUDA-graph capture sizes and its batch's range).
The solver finds the same pins.
In each engine, all arms (Table~\ref{tab:arms}) run the same version (\S\ref{sec:impl}) with FA3, CUDA graphs, prefix caching, chunked prefill and greedy decoding.

\paragraph{Workloads.}
W-A replays 600 requests from about 160 ShareGPT conversations~\citep{sharegpt2023} turn by turn.
Its prompts are short, and about 64\% of their tokens repeat the previous turn.
W-A exercises prefix caching and decode.
W-B sends 200 arXiv documents~\citep{cohan2018arxiv} of about 8k tokens with Poisson arrivals.
It mixes long prefills with decode under load.
W-B-c sends the first 64 of them in a closed loop to measure latency without queueing.
W-A and W-B can queue.
Running each of these three workloads at several loads gives nine workloads.
A1--A4 run W-A at 0.25--1.0$\lambda_{\max}$, B1--B3 run W-B at 0.33--1.0$\lambda_{\max}$, and C1, C2 run W-B-c at concurrency 1 and 8.
Here $\lambda_{\max}$ is Stock's maximum offline throughput in that engine.
In vLLM we serve A4, B3, C1 and C2.
On Qwen3-30B-A3B, C1 and C2 there send all 200 documents.
W-B and W-B-c run with a 32k-token context limit.
A workload tuple that is legal only at 32k serves all nine workloads at 32k.
Every Llama-3.1-70B arm runs GSM8K at 32k.

\paragraph{Tuples and metrics.}
\sys serves three kinds of tuple: the default tuple, a tuned tuple for each workload A1--C2 (\S\ref{sec:tuner}), and a workload tuple for each of W-A, W-B and W-B-c.
Each stays fixed while serving.
A workload tuple has the lowest mean cost over that workload's loads in an exhaustive table that predicts every legal tuple's GPU time on the nine workloads.
We report the p50, p90 and p99 of a request's latency per output token as the median of three runs.
A ratio is a latency over the same engine's Stock, at the geometric mean over the workloads A1--C2 that the engine serves.
We test batch invariance with a battery of 26 cells.
Each cell runs a request at temperature 0 in a reference batch and again in the same or a changed batch.
It compares the bits of the request's first 16 output logprobs.
This also catches a change in the logits that leaves the tokens alone.
Two cells rerun the same batch to check reproducibility, fourteen change the co-batched requests, four move the request to another row, and six change how much of its prompt hits the prefix cache.
Nine further strong cells place the request next to 1 to 32 short, long or mixed requests and move it within the batch.

\subsection{Solving and Tuning Costs}
\label{sec:eval-kernels}
\label{sec:eval-solver}

\paragraph{Ablating the solver's optimizations.}
On the main deployments the solver finds 3+3+0 pins (attention + all-reduce + MoE) on Llama-3.1-70B and 3+3+3 on Qwen3-30B-A3B.
Both engines get the same pins.
Each selection with pins needs a different part of the analysis.
With constant folding alone, FA3's cut has 12 pins instead of 3.
Without a cut, pinning every value the MoE's batch-dependent selection points read takes 10 pins instead of 3.
Each extra pin multiplies the tuner's search space by its number of candidates.
Without the analysis, FA3's selection has $2^{27}$ candidate assignments (Appendix~\ref{app:space}).
Without one of the simplifications, all but one of the all-reduce's legal assignments bypass the selection's own tuned table.
The GEMM selection has no cut.
The batch reaches neither of its two selection points, the kernel and the padding.

\paragraph{Solving and tuning time.}
Once per deployment, judging FA3's 7,168 candidate assignments takes 130--180\,s on 32 cores.
Repinning takes about 4\,ms.
With a five-minute budget per family, tuning one workload took 605--1,148\,s on Qwen3-30B-A3B and 690--1,006\,s on Llama-3.1-70B.
This includes worker startup.
For some workloads we also tuned with 10- or 30-minute budgets and kept the fastest tuple.

\subsection{Invariance and Serving}
\label{sec:eval-perf}

\begin{figure*}[t]
\centering
\includegraphics[width=\textwidth,trim=0 2.7pt 0 3.5pt,clip]{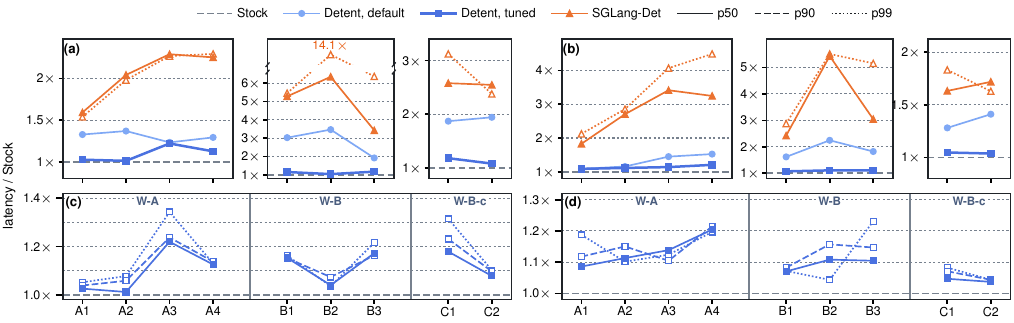}%
\caption{Latency per output token over Stock's in SGLang, median of three runs. (a, c)~Qwen3-30B-A3B, (b, d)~Llama-3.1-70B. A1--A4 (W-A), B1--B3 (W-B): loads up to Stock's peak. C1, C2 (W-B-c): one and eight in flight. (a, b)~p50, \sgdet also p99, each workload on its own scale. (c, d)~Tuned \sys at p50, p90 and p99: 1.01--1.22\tx at p50, 1.04--1.34\tx at p90 and p99.}
\label{fig:serving}
\end{figure*}
\begin{figure*}[t]
\centering
\includegraphics[width=\textwidth,trim=0 2.7pt 0 0.6pt,clip]{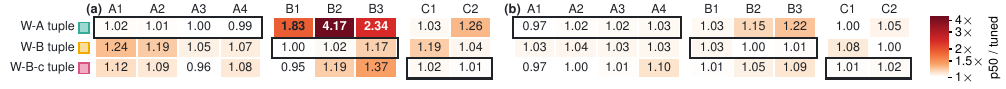}%
\caption{Rows: the workload tuples of W-A, W-B and W-B-c. Columns: the loads. Cells: the row's p50 over the p50 of the column's tuned tuple. (a)~Qwen3-30B-A3B, (b)~Llama-3.1-70B. Outlined: a tuple at its own workload's loads. On another workload's loads a tuple can take up to 4.17\tx (Qwen3-30B-A3B) and 1.22\tx (Llama-3.1-70B) as long.}
\label{fig:heatmap}
\end{figure*}
\begin{figure}[t]
\centering
\includegraphics[width=\columnwidth]{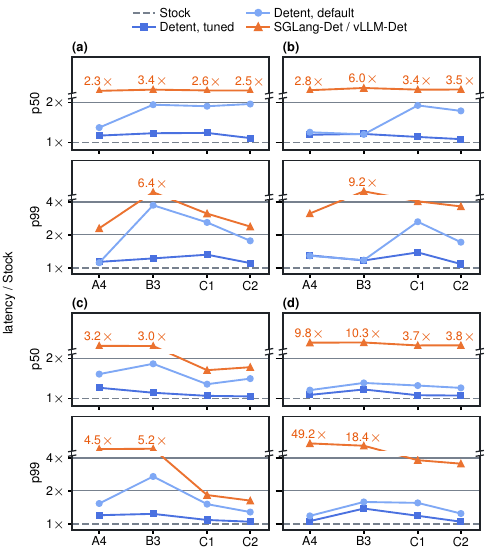}%
\caption{Latency per output token over Stock's at the loads vLLM serves, p50 and p99 (rows). SGLang (left), vLLM (right). (a, b)~Qwen3-30B-A3B, (c, d)~Llama-3.1-70B.}
\label{fig:xplat}
\end{figure}

\paragraph{Invariance.}
\sys, \sgdet and \vldet are batch-invariant on both models.
Stock fails 9 (Llama-3.1-70B) and 5 (Qwen3-30B-A3B) of 26 cells in SGLang, and 19 and 2 of the 24 and 23 cells that apply in vLLM.
With 16-token pages, the request's short prompt gets no prefix hit in the remaining cache-split cells.
\sys and the deterministic modes pass all nine strong cells.
Stock fails 4 to 7 of them.
Every KV-splitting tuple we served and ran the battery on also passes it.

\paragraph{Latency.}
Tuned per workload, \sys's p50 is 1.10\tx Stock's on Llama-3.1-70B and 1.11\tx on Qwen3-30B-A3B in SGLang (Figure~\ref{fig:serving}).
In vLLM it is 1.08\tx and 1.11\tx (Figure~\ref{fig:xplat}).
Across workloads its p50 is 1.01--1.22\tx and its geometric-mean p99 1.12--1.22\tx.
At the geometric mean, \sgdet is 2.4--2.6\tx slower than \sys at p50 and 2.8--3.0\tx at p99.
\vldet is 3.4--5.7\tx slower at p50 and 3.7--9.1\tx at p99.
On one workload their p50 reaches 6.1\tx and 9.2\tx \sys's.
Both values come from runs whose queue keeps growing.
We also serve each arm at 0.67 to 1.0 of its own maximum throughput so that its queue does not grow.
\vldet on Llama-3.1-70B's W-A and \sgdet on Qwen3-30B-A3B's W-B then still take 2.0--2.3\tx \sys's p50.

\paragraph{Ablating the tuner.}
Without tuning, the default tuple is up to 103\% (Llama-3.1-70B) and 1--233\% (Qwen3-30B-A3B) slower than the tuned tuple in SGLang (Figure~\ref{fig:serving}a,b).
In vLLM the gaps are up to 18\% and 72\%.
At p99 they reach 120\% and 273\% in SGLang and 31\% and 91\% in vLLM.
The tuples choose from the same kernels.
Tuning alone therefore accounts for 1.34\tx (Llama-3.1-70B) and 1.67\tx (Qwen3-30B-A3B) of \sgdet's gap at the geometric mean, and 1.13\tx and 1.31\tx of \vldet's.
Two things explain part of the remaining gap to Stock.
First, the rebuilt FA3 adds 3\,\textmu s per call, and its best split stride is within 1.06\tx of Stock's split.
Second, one tuple serves long prefills and small decode batches.

\paragraph{Ablating our kernels.}
Swapped into Stock, \sys's GEMM and all-reduce kernels add 0.6\% (Qwen3-30B-A3B) and 2.8\% (Llama-3.1-70B) GPU time per output token.
So they are not why \sys is close to Stock.
With Stock's FlashInfer all-reduce fusion on, \sys's decode step takes 0.996--1.103\tx Stock's.
Turning the fusion off does not weaken Stock (Appendix~\ref{app:ablation}).

\paragraph{Workload tuples.}
Stock adapts its selection to every batch.
A commitment tuple forbids this.
We adapt per workload instead.
Figure~\ref{fig:heatmap} shows what one tuple for all loads of W-A, W-B or W-B-c costs.
On its own workload, the tuple is within 2.7\% (Qwen3-30B-A3B) and 1.1\% (Llama-3.1-70B) of the tuned tuples at the geometric mean.
Another workload's tuple can be much slower.
On Qwen3-30B-A3B, the W-A tuple at B2 takes 4.17\tx as long as the tuned tuple and raises the median time to first token from 238\,ms to 4.75\,s.
On Llama-3.1-70B that cell takes 1.15\tx.
Off-diagonal cells are what traffic drift costs.
If traffic moves and the tuple stays, latency moves to that cell.

\subsection{Accuracy}
\label{sec:eval-acc}

\begin{figure*}[t]
\centering
\includegraphics[width=\textwidth]{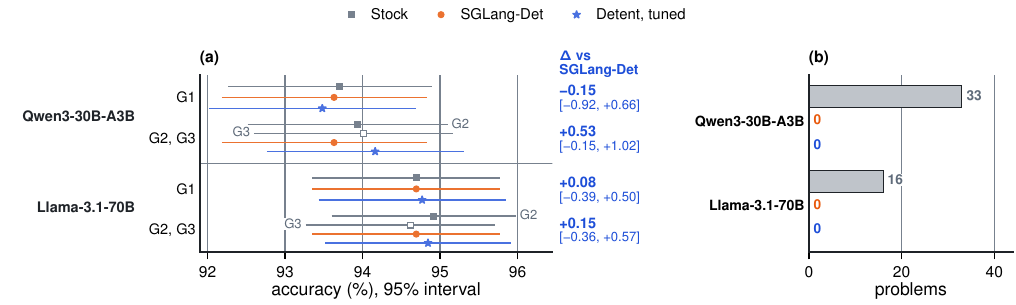}%
\caption{\sys keeps GSM8K accuracy. (a)~Accuracy with 95\% Wilson intervals. Right: \sys minus \sgdet with its 95\% interval. (b)~Answers that change across G1--G3 (\sys: G2 vs.\ G3, as its G1 tuple splits the KV).}
\label{fig:accuracy}
\end{figure*}

\begin{table}[!htb]
\caption{GSM8K accuracy (\%) on the 1,319 test problems, four comparisons. At G2 and G3, \sgdet's and \sys's outputs are identical, so the two load points form one comparison and only Stock has two accuracies (G2, G3). $\Delta$ is \sys minus \sgdet in percentage points with its 95\% interval, and $p$ the $p$-value of the exact McNemar test. Wilson 95\% intervals on the accuracies are about $\pm$1.3 percentage points.}
\label{tab:gsm8k}
\vskip 0.15in%
\centering
\scriptsize
\setlength{\tabcolsep}{2.6pt}
\begin{tabular}{@{}lccccc@{}}
\toprule
Point & Stock & \sgdet & \sys & $\Delta$ [95\%] & $p$ \\
\midrule
\multicolumn{6}{@{}l}{\textit{Qwen3-30B-A3B}} \\
G1 & 93.71 & 93.63 & 93.48 & $-$0.15 [$-$0.92, +0.66] & 0.85 \\
G2, G3 & 93.93, 94.01 & 93.63 & 94.16 & +0.53 [$-$0.15, +1.02] & 0.14 \\
\midrule
\multicolumn{6}{@{}l}{\textit{Llama-3.1-70B}} \\
G1 & 94.69 & 94.69 & 94.77 & +0.08 [$-$0.39, +0.50] & 1.00 \\
G2, G3 & 94.92, 94.62 & 94.69 & 94.84 & +0.15 [$-$0.36, +0.57] & 0.75 \\
\bottomrule
\end{tabular}
\end{table}

\sys runs different kernels from \sgdet, so its outputs can differ from \sgdet's.
They should be as accurate and should not depend on the load.
We serve GSM8K's 1,319 test problems~\citep{cobbe2021gsm8k} five-shot and greedily in SGLang at three load points: concurrency 8 (G1), concurrency 64 (G2) and all problems at once (G3, an overload point).
Each problem meets different batches at the three points.
An engine whose output depends on the batch can answer it differently.
We tune \sys on the training split at each point.
Its G1 tuple splits the KV, and its G2 and G3 tuples do not.
vLLM runs the same \sys, so we did not repeat these runs in vLLM.

\paragraph{\sys is as accurate as \sgdet} (Table~\ref{tab:gsm8k}, Figure~\ref{fig:accuracy}a).
\sys's accuracy minus \sgdet's is $-$0.15 to +0.53 percentage points (95\% intervals on the difference within $[-0.92, +1.02]$).
The exact McNemar test~\citep{mcnemar1947} gives $p \geq 0.14$.
The intervals on the accuracies in Figure~\ref{fig:accuracy}a are 95\% Wilson intervals~\citep{wilson1927}.
On the 1,209 corrected problems of GSM8K-Platinum~\citep{vendrow2025platinum}, accuracy is 96.11--96.77\% on Qwen3-30B-A3B and 96.86--97.11\% on Llama-3.1-70B.
There \sys stays within $[-0.93, +0.96]$ points of \sgdet (McNemar's $p$ 0.18--1).

\paragraph{Only Stock's answers depend on the load} (Figure~\ref{fig:accuracy}b).
Across the three points, Stock's text changes on 790 (Qwen3-30B-A3B) and 484 (Llama-3.1-70B) of the 1,319 problems, and its answer on 33 and 16.
\sgdet's text never changes.
\sys serves G2 and G3 with different tuples but gives identical text, so \sgdet and \sys each form one comparison across G2 and G3.
\sys's G1 text can differ from its G2 text because its G1 tuple splits the KV and so sums in a different order.
Its answers follow its tuple, not the load.
At G1 and G2, \sys's p50 latency per output token is 1.006--1.057\tx Stock's, and \sgdet's is 1.38--1.76\tx \sys's.
At G3, \sys's throughput is 0.945\tx and 0.934\tx Stock's.
So \sys gains its speed without losing accuracy.

\subsection{Comparison with LLM-42}
\label{sec:eval-llm42}

\begin{figure*}[t]
\centering
\includegraphics[width=\textwidth]{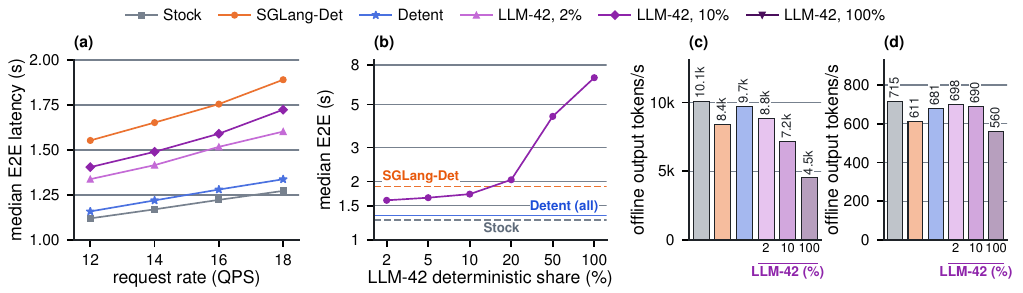}%
\caption{\sys makes every request deterministic at 3.5--5.1\% above Stock's median latency. LLM-42 costs more even when only 2\% of requests are deterministic. Llama-3.1-8B on one H100, ShareGPT online, median of three servers per point. (a)~Online, median end-to-end latency against request rate. LLM-42 at 100\% (4.6--6.9\,s) is off this scale. (b)~Online at 18 requests per second, LLM-42 as its deterministic share grows. (c,~d)~Offline output throughput on ShareGPT and arXiv, one run per configuration. LLM-42 runs on its port to our SGLang base. \sys's tuner judged a tie on this workload and committed the default tuple.}
\label{fig:llm42}
\end{figure*}

\begin{table}[t]
\caption{Online: ported LLM-42's median end-to-end latency over \sys's on Llama-3.1-8B, at four request rates and six deterministic shares.}
\label{tab:llm42}
\vskip 0.15in
\centering
\scriptsize
\setlength{\tabcolsep}{4pt}
\begin{tabular}{@{}lcccc@{}}
\toprule
Deterministic & 12 QPS & 14 QPS & 16 QPS & 18 QPS \\
\midrule
2\% & 1.155 & 1.161 & 1.185 & 1.198 \\
5\% & 1.187 & 1.191 & 1.212 & 1.236 \\
10\% & 1.212 & 1.221 & 1.242 & 1.289 \\
20\% & 1.265 & 1.302 & 1.379 & 1.530 \\
50\% & 2.023 & 2.408 & 2.842 & 3.249 \\
100\% & 3.946 & 4.322 & 4.756 & 5.147 \\
\bottomrule
\end{tabular}
\end{table}

\begin{table}[t]
\caption{Offline: \sys's output throughput over ported LLM-42's on Llama-3.1-8B, at six deterministic shares.}
\label{tab:llm42off}
\vskip 0.15in
\centering
\scriptsize
\setlength{\tabcolsep}{4pt}
\begin{tabular}{@{}lcc@{}}
\toprule
Deterministic & ShareGPT & arXiv \\
\midrule
2\% & 1.098 & 0.975 \\
5\% & 1.224 & 0.982 \\
10\% & 1.357 & 0.987 \\
20\% & 1.471 & 0.996 \\
50\% & 1.727 & 1.063 \\
100\% & 2.147 & 1.217 \\
\bottomrule
\end{tabular}
\end{table}

LLM-42~\citep{llm42} makes a chosen share of requests deterministic without changing the kernels.
It verifies fast-path tokens and rolls back mismatches.
We port it to our SGLang base and reproduce its single-GPU setting on one H100 SXM.
That setting runs Llama-3.1-8B on ShareGPT at 12 to 18 requests per second (QPS) with 2 to 100\% of requests deterministic.
LLM-42 used an H100 PCIe.
\sys's tuner judged a tie on this workload and committed the default tuple.
With prefix caching on, \sys is batch-invariant while Stock fails 14 of the 26 cells.

\paragraph{Online.} With 4,096 requests and three independent servers per point, \sys's median end-to-end latency is 3.5--5.1\% above Stock's, while \sgdet's is 39--49\% above.
Ported LLM-42 is slower than \sys at every rate and deterministic share (Figure~\ref{fig:llm42}a,b and Table~\ref{tab:llm42}).
At 2\%, LLM-42 makes only 2\% of requests deterministic and \sys makes all of them.
Even so, LLM-42's median latency there is 1.15--1.20\tx \sys's.
No deterministic arm's output changed at the token level: 0 of 45,184 comparisons for \sys and 0 of 87,916 for LLM-42's deterministic requests.

\paragraph{Offline.} \sys's throughput is 0.964\tx (ShareGPT) and 0.952\tx (arXiv) Stock's.
It is 1.16\tx and 1.11\tx \sgdet's.
On ShareGPT, it is 1.10--2.15\tx ported LLM-42's (Table~\ref{tab:llm42off}).
On arXiv, LLM-42 at 20\% deterministic or less has 0.4--2.6\% higher throughput than \sys.

\paragraph{Fidelity of the port.}
On the official fork's online grid (12 and 18 requests per second at 2, 10 and 100\%), the port is within 1.1\% of official LLM-42.
The pre-registered offline check failed by 44\%.
It divided each stack's LLM-42 throughput by that of the stack's own SGLang deterministic mode.
That mode runs 1.76\tx as fast on our SGLang base as on the official fork's older one.
We corrected the check after seeing the data.
It now compares LLM-42 throughput directly and passes within its 10\% tolerance.
The port's throughput is 0.987\tx official LLM-42's.

\subsection{On-Policy RL Rollouts and Training}
\label{sec:eval-rl}

\begin{figure*}[t]
\centering
\includegraphics[width=\textwidth,trim=0 1.3pt 0 3.5pt,clip]{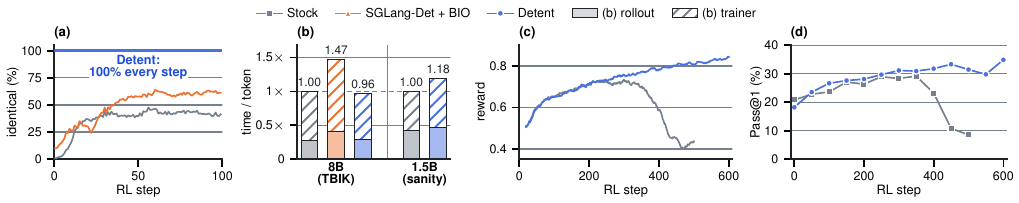}%
\caption{\sys's trainer matches the rollout's bits at near-stock cost. Its GRPO run does not collapse. (a)~Bit-identical tokens and (b)~time per token over Stock's, TBIK's protocol on Llama-3.1-8B. Sanity test on DeepSeek-R1-Distill-Qwen-1.5B, (c)~training reward (20-step mean), (d)~AIME 2024.}
\label{fig:rl}
\end{figure*}
\begin{figure*}[t]
\centering
\includegraphics[width=\textwidth,trim=0 2.2pt 0 3.5pt,clip]{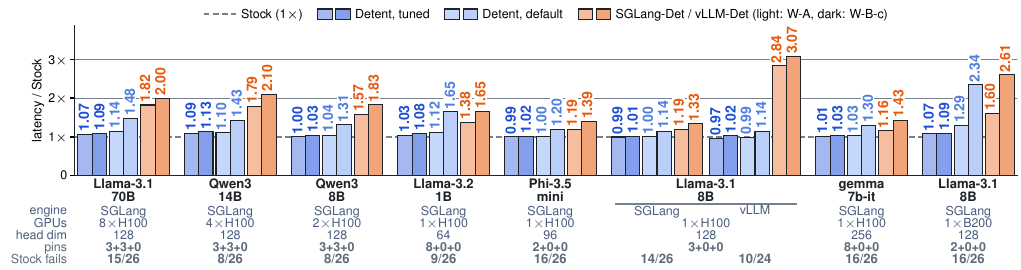}%
\caption{On eight more deployments and a B200, solved from specs alone, \sys is batch-invariant. Bars: closed-loop median at W-A and W-B-c. Pins: attention + all-reduce + MoE. Stock fails: battery cells where Stock's bits change.}
\label{fig:general}
\end{figure*}

For RL, we commit seeded temperature sampling and refuse any other sampling.
\sys's trainer runs the serving kernels under the same tuple (\S\ref{sec:overview-rl}).
As in TBIK~\citep{tbik2026}, we compare it with Stock and with \sgdet plus a transformers trainer on batch-invariant operators (BIO).
Stock trains with transformers.

\begin{table}[t]
\caption{Seeded rollouts. 512/512: bit-identical replays, rescorings. Ratios: \sys's throughput over Stock's and \sgdet's.}
\label{tab:rlroll}
\vskip 0.15in%
\centering
\scriptsize
\setlength{\tabcolsep}{2.3pt}
\begin{tabular}{@{}lccccc@{}}
\toprule
\multirow{2}{*}{Model, H100s} & \multicolumn{4}{c}{\sys} & \sgdet \\
\cmidrule(lr){2-5}
 & BI & Replay, rescoring & / Stock & / \sgdet & BI, mixed \\
\midrule
Llama-3.1-8B, 1 & yes & 512/512 & 0.917\tx & 1.15\tx & no \\
Qwen3-30B-A3B, 4 & yes & 512/512 & 0.837\tx & 1.27\tx & no \\
Llama-3.1-70B, 4 & yes & 512/512 & 0.872\tx & 1.35\tx & no \\
\bottomrule
\end{tabular}
\end{table}

\paragraph{Seeded rollouts.}
Table~\ref{tab:rlroll} gives \sys's seeded rollouts on Llama-3.1-8B (one H100) and on Qwen3-30B-A3B and Llama-3.1-70B (four H100s each).
Its seeded sampling is batch-invariant.
All 512 GSM8K~\citep{cobbe2021gsm8k} rollouts at temperature 1 replay bit for bit under other batches and cache splits and match their rescoring.
At 1,024 output tokens, its throughput is 0.84--0.92\tx Stock's and 1.15--1.35\tx \sgdet's.
\sgdet's seeded rollouts change in all 20 cells where one co-batched request is greedy or uses top-p.
They hold when every request samples with temperature.

\paragraph{Trainer agreement.}
We use TBIK's on-policy protocol on Llama-3.1-8B (GSM8K, full fine-tuning, 100 steps) but lower the learning rate from TBIK's $10^{-5}$ to $10^{-6}$.
\sys's trainer matches the rollout's bits on every token of every step, so KL is 0.
Stock matches 1.0--47.7\% of the tokens and \sgdet with BIO 9.9--63.9\% (Figure~\ref{fig:rl}a).
Per generated token, \sys's rollout takes 1.03\tx Stock's time and its whole step stays within 4\% of Stock's, while \sgdet with BIO takes 1.47\tx (Figure~\ref{fig:rl}b).
Its median step takes 1.09\tx Stock's because it generates 11\% more tokens.

\paragraph{BF16 GRPO stability.}
In the sanity test of \citet{qi2025fp16}, BF16 GRPO on DeepSeek-R1-Distill-Qwen-1.5B~\citep{guo2025deepseekr1} collapses.
We run that recipe in \sys with one GPU serving and one training.
Stock's reward falls from 0.733 (step 299) to 0.436 (step 500), and its AIME 2024~\citep{aime2024} Pass@1 from 29.2\% to 8.5\% (Figure~\ref{fig:rl}c,d).
\sys keeps KL at 0 through step 600, and its reward reaches 0.843.
It takes 1.18\tx Stock's time per token, most of it in training.

\subsection{Ablation Across Deployments}
\label{sec:eval-port}

Figure~\ref{fig:general} adds eight deployments solved from specs alone.
They include Llama-3.1-70B at TP=8, Phi-3.5-mini~\citep{abdin2024phi3}, gemma-7b-it~\citep{gemma2024} and Llama-3.1-8B in both engines.
With the main deployments and a B200, they span two engines, two GPU generations, tensor-parallel degrees 1, 2, 4 and 8, head dimensions 64 to 256, and dense and MoE models.
Each ablates one part of the main deployments.
\sys is batch-invariant on all eight, while Stock changes outputs in 8 to 16 cells.
On these eight, \S\ref{sec:eval-perf}'s tuner ablation repeats.
The default tuple, which never splits the KV, takes 0.99--1.14\tx Stock's latency on W-A and 1.14--1.65\tx on W-B-c.
Tuning each workload brings this to 0.97--1.09\tx and 1.01--1.13\tx, which is 1.14--3.01\tx faster than the deterministic modes.
On a B200 with Llama-3.1-8B, \sys and \sgdet pass 26 cells while Stock fails 16.
There, the default tuple takes 1.29\tx (W-A) and 2.34\tx (W-B-c) Stock's latency.
The tuned tuples take 1.07\tx and 1.09\tx, 1.49\tx and 2.38\tx faster than \sgdet.

\section{Related Work}

\paragraph{Batch-invariant kernels.}
\sys takes batch-invariant kernels as given and makes the selection among them batch-invariant.
Thinking Machines wrote batch-invariant kernels with one matrix-multiplication configuration and a fixed attention split size~\citep{thinkingmachines2025}.
SGLang's deterministic mode builds on them and runs FA3 with one split~\citep{sglangdet2025}.
So does vLLM's batch-invariant mode~\citep{vllmbi2026}.
TBIK fixes one reduction tree, so outputs also match across TP degrees~\citep{tbik2026}.

\paragraph{Controlling what kernels see.}
LLM-42 decodes on the fast path, replays tokens under a fixed reduction schedule and rolls back mismatches (\citealp{llm42}; \S\ref{sec:eval-llm42}).
CoRun prefills requests alone and pads decode batches to a fixed row count, with split-KV off~\citep{corun2026}.
Both control the batches that stock kernels see. \sys instead fixes what kernel selection reads. This lets it keep split-KV and reuse prefixes across turns.

\paragraph{RL rollout and training.}
Rollout and training engines disagree on token probabilities~\citep{yao2025mismatch}.
\citet{qi2025fp16} switch to FP16 to narrow the gap.
Thinking Machines, Miles, VeXact and TBIK make the two bit-identical with batch-invariant kernels~\citep{thinkingmachines2025,miles2025,zhong2026vexact,tbik2026}. Our trainer gets the same result from pinned kernel selection at near-stock speed.

\paragraph{Reproducibility and verification.}
Hawkeye replays GPU matrix multiplication on a CPU~\citep{badash2026hawkeye}, and EigenAI audits inference by re-execution~\citep{eigenai2026}.
TOPLOC and DiFR verify inference despite nondeterminism~\citep{ong2025toploc,karvonen2025difr}, and Vosti verifies a single-GPU engine~\citep{qin2026vosti}.
\sys makes multi-GPU engines like vLLM and SGLang batch-invariant.

\paragraph{Kernel selection and autotuning.}
Engines adapt kernels to each step's batch~\citep{kwon2023vllm,zheng2024sglang,ye2025flashinfer,gond2026tokenweave}.
Triton's autotuner re-measures when its arguments change~\citep{tillet2019triton,tritonautotune}.
\sys tunes its pins once per workload before serving.

\paragraph{Program analysis.}
The cut enforces the information-flow policy~\citep{denning1976} that batch state must not reach kernel selection.
Quantitative information flow~\citep{mccamant2008qif} and Swift~\citep{chong2007swift} also use maximum flows.
\sys uses partial evaluation~\citep{jones1993pe}, slicing~\citep{weiser1984} and value range analysis~\citep{harrison1977vra}.
VRA applies it to pins.

\raggedbottom%
\section{Conclusion}

\sys makes LLM serving engines such as vLLM and SGLang batch-invariant while keeping their fast kernels.
It pins the few values through which the batch reaches kernel selection and tunes them per workload, so latency per output token stays at 1.04--1.17\tx Stock's.
The engines' deterministic modes take 1.67--10.3\tx.
A trainer on the same kernels and tuple matches every rollout logprob bit for bit at near-stock speed, keeping KL at 0.
BF16 GRPO runs stay stable with \sys where Stock's collapses.
Because selection then depends only on the deployment, the tuple and the request, anyone with the same deployment and tuple can re-execute a request and check its output bit for bit.

\needspace{4\baselineskip}%
\paragraph{Limits and next steps.}
\begin{itemize}[leftmargin=1em, itemsep=0pt, topsep=0pt, parsep=0pt]
\item \emph{Scope.} Outputs match only within one deployment, model and commitment tuple. Other TP degrees also need TP-invariant kernels like TBIK's.
\item \emph{Writing \selir.} Users need to write each selection in \selir. We plan LLM-assisted transcription and native-code selectors.
\item \emph{Traffic drift.} Speed depends on how well the tuned workloads match traffic. Re-tuning must switch tuples where no batch mixes them.
\item \emph{Quantization.} Dynamic per-tensor FP8 activation scales span the batch. Batch-invariant FP8 serving needs per-token or static scales.
\item \emph{Agent RL.} Agentic rollouts grow long contexts and end on-policy steps with few requests, where split-KV matters most. \sys is one of the few batch-invariant systems that keep it, so agent RL is our next target.
\end{itemize}

\bibliography{references}
\bibliographystyle{mlsys2025}

\clearpage
\appendix
\flushbottom

\section{The Deterministic Modes' Switches}
\label{app:det}
We turn on SGLang's deterministic mode with \code{--enable-}\allowbreak\code{deterministic-}\allowbreak\code{inference} and vLLM's batch-invariant mode with \code{VLLM\_\penalty200{}BATCH\_\penalty200{}INVARIANT=1}. Table~\ref{tab:det} lists what each mode changes in the versions we run. Figure~\ref{fig:hardness}d also shows SGLang's changes. Both modes keep one conservative kernel per operator.

\begin{table*}[!htb]
\caption{What the engines' deterministic modes fix.}
\label{tab:det}
\vskip 0.15in%
\centering
\scriptsize
\setlength{\tabcolsep}{3pt}
\begin{tabular}{@{}>{\raggedright\arraybackslash}p{0.15\textwidth}>{\raggedright\arraybackslash}p{0.25\textwidth}>{\raggedright\arraybackslash}p{0.27\textwidth}>{\raggedright\arraybackslash}p{0.29\textwidth}@{}}
\toprule
Operator & Stock & \sgdet & \vldet \\
\midrule
FA3 & split count from a batch-wide heuristic & one split & one split; ahead-of-time scheduler and cascade attention off \\
All-reduce (TP$>$1) & custom, symmetric-memory or NCCL kernels by message size & NCCL tree with fixed channels & NCCL tree, one channel; custom and symmetric-memory all-reduce off \\
Fused MoE & tuned table, nearest token count & one hard-coded configuration & one hard-coded configuration; kernels without batch-invariant support refused \\
GEMM & cuBLAS, kernel picked by shape & batch-invariant \code{mm}, \code{addmm} on DeepGEMM & Triton persistent matmul for linear layers; cuBLAS without split-K \\
Other row-local operators & native kernels & batch-invariant \code{log\_softmax}, \code{mean}, \code{rms\_norm} & batch-invariant \code{log\_softmax}, \code{softmax}, \code{mean}, \code{rms\_norm} \\
Sampling & default backend & PyTorch backend & unchanged \\
\bottomrule
\end{tabular}
\end{table*}

\section{Counting the Selection Space}
\label{app:space}
These are the counts behind Figure~\ref{fig:hardness}c, for the FA3 and SGLang versions we evaluate.
FA3's \code{run\_flash\_fwd} has 19 template parameters, 13 of them Boolean.
On sm90, the product of the numbers of values they can take is 737,280.
A full build compiles 4,720 forward kernels, 3,280 of them for sm90.
A split kernel takes any split count from 2 to 256.
On sm90 this gives 332,464 kernel configurations, 18,472 of them at head dimension 128 in bf16.
One tuning run on Qwen3-30B-A3B at TP=4 spent 22.4\,s per measured point on average. At that rate, measuring the 18,472 configurations at head dimension 128 in bf16 once takes 4.8 days.
Without the analysis, FA3's selection has $2^{27}$ candidate assignments.
SGLang's fused-MoE tuning grid has 1,920 configurations for each of 18 token counts.
One tuned table takes 34,560 runs.

\section{Ablation of Stock's Kernels}
\label{app:ablation}
\begin{figure*}[!t]
\centering
\includegraphics[width=\textwidth]{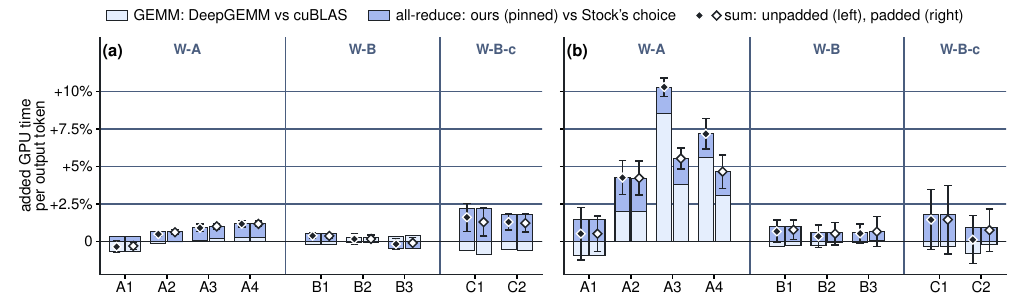}%
\caption{What \sys's GEMM and all-reduce kernels add to Stock's GPU time per output token at each workload, split by family. Left bar: the GEMM read unpadded. Right bar: with the padding that workload's tuple fixes. The padding leaves the all-reduce untouched. A1 and A4 in (a) and A1 and C1 in (b) are served unpadded, so their bars are identical. (a)~Qwen3-30B-A3B, +0.6\% and +0.6\% on average; (b)~Llama-3.1-70B, +2.8\% and +2.1\%. Diamonds: the sums, with twice the measurement noise of the unpadded reading.}
\label{fig:ablation}
\end{figure*}
Stock and \sys differ in more than kernel selection. They use different matrix-multiplication and all-reduce kernels. Every arm runs with FlashInfer's all-reduce fusion off (Table~\ref{tab:arms}). This appendix ablates both differences. It checks that neither weakens Stock, that is, that \sys does not come close to Stock by giving itself faster kernels or by taking a feature away from Stock.

\begin{table}[!htb]
\caption{Ablation of Stock's kernels. Top: \sys's kernels over Stock's in GPU time (geometric mean and range over the workloads), and the GPU time they add to Stock's GPU time per output token (mean and range). Middle: the all-reduce row broken down by the algorithm the tuple fixes and by the algorithm Stock picks for each message (in parentheses: share of \sys's extra all-reduce time). Bottom: \sys's decode step over Stock's when Stock runs FlashInfer's all-reduce fusion.}
\label{tab:ablation}
\vskip 0.15in%
\centering
\scriptsize
\setlength{\tabcolsep}{3pt}
\begin{tabular}{@{}lcc@{}}
\toprule
Measure & Qwen3-30B-A3B & Llama-3.1-70B \\
\midrule
\multicolumn{3}{@{}l}{\textit{Kernel swap: \textup{\sys}'s kernel over Stock's}} \\
GEMM: DeepGEMM / cuBLAS & 0.99 (0.93--1.05) & 1.02 (0.99--1.15) \\
\quad with the tuple's padding & 1.00 (0.93--1.06) & 1.01 (0.99--1.06) \\
All-reduce: ours / Stock's pick & 1.10 (0.97--1.29) & 1.16 (1.04--1.37) \\
Added to Stock's GPU time & +0.6\% ($-$0.3 to +1.6) & +2.8\% (+0.1 to +10.3) \\
\quad with the tuple's padding & +0.6\% ($-$0.3 to +1.3) & +2.1\% (+0.5 to +5.5) \\
Twice the noise & 0.5\% & 1.4\% \\
\midrule
\multicolumn{3}{@{}l}{\textit{All-reduce by the tuple's workspace and Stock's pick}} \\
push, Stock one-shot push & 1.003 (1\%) & 1.003 (0\%) \\
push, Stock two-shot pull & 1.54 (15\%) & 1.40 (2\%) \\
push, Stock NCCL & 2.03 (61\%) & 2.03 (1\%) \\
pull, Stock one-shot push & 1.68 (39\%) & 1.72 (84\%) \\
pull, Stock two-shot pull & 1.10 (2\%) & 1.21 (6\%) \\
pull, Stock NCCL & 0.94 ($-$18\%) & 1.01 (6\%) \\
\midrule
\multicolumn{3}{@{}l}{\textit{Stock with FlashInfer's all-reduce fusion: decode step, \textup{\sys}{} / Stock}} \\
1 request per step & 1.103 & 1.009 \\
32 requests per step & 0.996 & 1.046 \\
\bottomrule
\end{tabular}
\end{table}

\textbf{Matrix-multiplication and all-reduce kernels.} We record each workload's GEMM and all-reduce calls in SGLang while profiling. We time them once on Stock's kernels and once on \sys's (Table~\ref{tab:ablation}, Figure~\ref{fig:ablation}). \sys's kernels are the slower ones. They add 0.6\% (Qwen3-30B-A3B) and 2.8\% (Llama-3.1-70B) to Stock's GPU time per output token. This is more than twice the measurement noise, so Stock keeps the faster kernels. The GEMM row runs DeepGEMM unpadded, so it weighs the two libraries and not \sys's padding knob. Read with the padding each tuple fixes, the two kernels add 0.6\% and 2.1\%. \sys's are still the slower ones. In most cases, most of the difference is in the all-reduce and comes from fixing one algorithm per tuple. When both run the one-shot push, \sys and Stock take the same time (1.003\tx). When both run the two-shot pull, \sys's own two-shot kernel takes 1.10--1.21\tx Stock's.

\textbf{FlashInfer's all-reduce fusion.} The fusion does little for Stock, so turning it off does not weaken Stock. On Qwen3-30B-A3B, Stock without the fusion is 1.3--2.1\% faster at a decode step with one request and 11--13\% faster on a 1,447-token prefill. From 8 to 256 requests, the difference on this model is within run-to-run noise. With the fusion on in Stock, \sys's decode step takes 0.996--1.103\tx Stock's on both models (Table~\ref{tab:ablation}, bottom).

\end{document}